# Identifying Metaphoric Antonyms in a Corpus Analysis of Finance Articles

**Aaron Gerow (gerowa@tcd.ie)**
School of Computer Science & Statistics, Trinity College Dublin
College Green, Dublin 2, Ireland

**Mark T. Keane (mark.keane@ucd.ie)**
School of Computer Science & Informatics, University College Dublin
Belfield, Dublin 4, Ireland

**Abstract**

Using a corpus of 17,000+ financial news reports (involving over 10M words), we perform an analysis of the argument-distributions of the UP and DOWN verbs used to describe movements of indices, stocks and shares. In Study 1 participants identified antonyms of these verbs in a free-response task and a matching task from which the most commonly identified antonyms were compiled. In Study 2, we determined whether the argument-distributions for the verbs in these antonym-pairs were sufficiently similar to predict the most frequently-identified antonym. Cosine similarity correlates moderately with the proportions of antonym-pairs identified by people ($r = 0.31$). More impressively, 87% of the time the most frequently-identified antonym is either the first- or second-most similar pair in the set of alternatives. The implications of these results for distributional approaches to determining metaphoric knowledge are discussed.

**Keywords**: Metaphor; corpus analysis; word meaning; semantics; experimental linguistics; grounding.

## Introduction

In recent years, significant progress has been made in deriving meaning from statistical analyses of distributions of words (Gerow & Keane, 2011a; Landauer & Dumais, 1997; Michel et al., 2010; Turney & Pantel, 2010). This distributional approach to meaning takes the view that words that occur in similar contexts tend to have similar meanings (cf. Wittgenstein, 1953) and that by analysing word usage we get at their meaning. For example, the word co-occurrence statistics derived in Latent Semantic Analysis (LSA) seem to tell us about the structure of the lexicon, as they are good predictors of reaction times in lexical decision tasks (Landauer & Dumais, 1997). More generally, it has been suggested that significant insights into human culture and behaviour can be derived from analysing very large corpora, like the Google Books repository (Michel et al., 2010). In this paper, we apply similar distributional analyses to understand metaphorically-structured knowledge underlying the antonyms between "UP" and "DOWN" verbs from a corpus of financial news reports. (see Gerow & Keane, 2011b, for an analysis of metaphor hierarchies in the same data.)

Lakoff (1992; Lakoff & Johnson, 1980) have argued that our understanding of many concepts, such as emotions and mental states, are grounded in a few ubiquitous metaphors. The spatial metaphors that structure emotional states – HAPPINESS IS UP and SADNESS IS DOWN – are found in almost all languages. Similar spatial metaphors, of the kind we examine here, seem to ground many stock-market reports. Accounts of index, stock-market, and share movements tend to converge around metaphors of *rising* and *falling*, *attack* and *retreat*, *gain* and *loss*. These concepts appear to be grounded by core metaphors, with an antonymic relationship to one another, that could be glossed as GOOD IS UP and BAD IS DOWN. Lakoff and Johnson (1980) have pointed to this UP-DOWN metaphor opposition as underlying accounts of wealth (WEALTH IS UP as in *high class*), the rise and fall of numbers (MORE IS UP; LESS IS DOWN) and changes in quantity (CHANGE IN QUANTITY IS WAR as in *retreating profits* and *defensive trades*).

In the present paper, we look at the distributive structure of these verbs' arguments to determine whether there is empirical support for metaphoric opposites. Specifically, we try to determine whether the antonyms identified by participants in a psychological study can be shown to meaningfully correspond to a computational analysis of the argument-distributions in our corpus.

## The Corpus

In January, 2010, we carried out automated web searches that selected all articles referring to the three major world stock indices (Dow Jones, FTSE 100, and NIKKEI 225) from three websites: the *New York Times* (NYT, www.nyt.com), the *Financial Times* (FT, www.ft.com) and the *British Broadcasting Corporation* (BBC, www.bbc.co.uk). These searches harvested 17,713 articles containing 10,418,266 words covering a 4-year period: January 1st, 2006 to January 1st, 2010. The by-source breakdown was FT (13,286), NYT (2,425), and BBC (2,002). The by-year breakdown was 2006 (3,869), 2007 (4,704), 2008 (5,044), 2009 (3,960), and 2010 (136). The corpus included editorials, market reports, popular pieces, and technical exposés. These three resources were chosen because they are in English and have a wide-circulation and online availability. The Financial Times made up the majority of the articles; however, the spread was actually much wider as many articles were syndicated from the Associated Press, Reuters, Bloomberg News, and Agence France-Presse. The uniqueness of the articles in the database was ensured by keying them on their first 50 characters.

Once retrieved, the articles were stripped of HTML, converted to UTF-8, and shallow-parsed to extract phrasal structure using a modified version of the Apple Pie Parser (Sekine, 1997). Each article was stored in a relational database with sentential parses of embedded noun- and verb-phrases. Sketch Engine was used to lemmatise and tag the corpus (Kilgarriff et al., 2004). Sketch Engine is a web-based, corpus-analysis tool that lemmatises and tags customised corpora with part-of-speech tags using the TreeTagger schema (Schmid, 1994). A lemma is a singular, part-of-speech token (e.g., verb or noun) that includes all tenses, declensions, and pluralizations of a given word. For example, the one verb lemma "fall" includes instances such as "fall", "fell" and "falls", whereas the noun lemma "fall" includes "a fall" and "three falls". Sketch Engine provides so-called "sketches" of individual lemmas. For example, the sketch for fall-n (the word "fall" as a noun) is different from the sketch for fall-v ("fall" as a verb.) With some lemmas, the differences marked by part-of-speech are large, such as with store-n compared to store-v. These sketches facilitated the statistical analysis of the most common arguments of verbs. For example, one of the most common verbs in the corpus was "fall," which took a range of arguments with different frequencies (e.g., "DJI", "stocks", "unemployment"). Throughout this paper, when we refer verbs we take this to mean verb lemmas.

Table 1: The percentage of the argument-distributions of *rise* and *fall* for their 10 most frequent arguments.

| *Rise* (Arg) | % of Total (Freq) | *Fall* (Arg) | % of Total (Freq) |
|---|---|---|---|
| index | 7.39% | index | 6.97% |
| share | 5.67% | share | 6.41% |
| point | 4.83% | point | 3.75% |
| percent | 2.90% | percent | 2.97% |
| price | 2.43% | price | 2.83% |
| stock | 2.00% | stock | 2.78% |
| yield | 1.90% | yield | 1.77% |
| cent | 1.31% | cent | 1.34% |
| profit | 0.91% | profit | 1.34% |
| rate | 0.90% | rate | 1.24% |

($N_{rise}$ = 39,261; $N_{fall}$ = 39,230).

## Metaphoric Antonyms

From a distributional perspective, the arguments of a verb and its antonym (like *rise* and *fall*) should have a definite structure that identifies their relationship to one another. That is, the frequency distribution of the arguments taken by *rise* should have a lot in common with the argument-distribution of its antonym *fall* (see Table 1). Furthermore, if we look at other less-strongly-paired antonyms, like *rise-lower* or *rise-decrease,* then the similarity in their argument distributions should be less. Specifically, we should find that a computational measure of similarity, such as cosine similarity, between the words' argument-distributions should be predictive people's choice of antonyms.

Within a larger body of work on automated semantic tagging and semantic parsing, some work has focused on automating the generation of semantically resolute phrases (Brown et al., 2005). Online lexicons, such as WordNet and LSA, have been used to generate and resolve analogies by modelling synonymy (Turney, 2006; Veale, 2004). Such work approaches semantics, and specifically antonymy, between words and phrases, but avoids conceptual metaphors. Lakoff (1992) offers a cognitive theory of metaphor, one in which linguistic metaphors are related, but distinct, from the metaphoric concepts they structure. Deignan (2005) offers a bridge between concept and language, by proposing a cline between metonymy (part-whole relationship) and metaphor. Deignan's link from metonymy to metaphor is a good example of a corpus-based approach to metaphor because it preserves the cognitive structures proposed by Lakoff, while making the link between semantics (words) and metaphor (thought) explicit. Here, we explore this link with regard to antonyms.

In this article, we report two studies examining these issues. Study 1 was a study of participants' identification of antonyms in two distinct tasks: a free-generation task (where one is given *rise* and asked for its opposite) and a match-the-opposite task (where one is asked to match *rise* to its opposite in a set of words). The word-sets were drawn from the above corpus and consisted of a set of positive, UP verbs (e.g., *rise, soar, rally*) and more negative, DOWN verbs (e.g., *fall, lose, dip*; see Table 2). Study 2 examined the argument-distributions of the antonym-pairs chosen by participants in Study 1 to see if they were, in any way, predictive of the choices made. To anticipate our findings, we find that argument distributions correlate moderately with the frequencies of antonym choices made by people. Furthermore, in the majority of cases, the most similar distribution for an antonym pair corresponds to the pair most-frequently chosen by people.

Table 2: The UP and DOWN verb used in studies.

| *UP-verbs* occurrences (% corpus[*]) | | *DOWN-verbs* occurrences (% corpus[*]) | |
|---|---|---|---|
| rise | 29,261 (4.20%) | fall | 39,230 (4.20%) |
| gain | 13,134 (1.40%) | lose | 12,298 (1.30%) |
| increase | 6,158 (0.67%) | decrease | 123 (0.01%) |
| climb | 5,631 (0.60%) | tumble | 2,135 (0.23%) |
| jump | 4,960 (0.53%) | slip | 3,336 (0.36%) |
| rally | 4,190 (0.45%) | retreat | 1,474 (0.20%) |
| advance | 2,385 (0.26%) | slide | 2,777 (0.30%) |
| surge | 2,313 (0.25%) | plunge | 1,592 (0.17%) |
| recover | 2,165 (0.23%) | worsen | 500 (0.05%) |
| soar | 1,649 (0.18%) | plummet | 443 (0.05%) |
| rebound | 1,220 (0.13%) | dip | 1,322 (0.14%) |
| alleviate | 134 (0.01%) | decline | 3,672 (0.39%) |
| elevate | 52 (0.01%) | drop | 8,377 (0.90%) |
| strong | 718 (0.07%) | weak | 1222 (0.13%) |
| ease | 2,243 (0.35%) | sink | 1,339 (0.14%) |

Table 3: The percentages of antonym-pairs identified in the two tasks (*T1* and *T2*) of Study 1 and their cosine similarity scores (*Sim*). *Total %* is the mean percent occurrence across both tasks; bold words were only generated in the free-response task (*T1*).

| Antonym pair (prompt-response) | Task 1 % | Task 2 % | Total % | Sim |
|---|---|---|---|---|
| advance-climb | 7% | 0% | 4% | **0.98** |
| advance-leave | 7% | 0% | 4% | 0.62 |
| alleviate-worsen | 40% | 0% | 20% | **0.86** |
| climb-fall | 46% | 17% | 31% | **0.98** |
| climb-plunge | 0% | 8% | 4% | 0.97 |
| decline-rise | 20% | 33% | 27% | **0.93** |
| decline-incline | 30% | 0% | 13% | 0.42 |
| drop-rise | 10% | 42% | 26% | **0.99** |
| drop-climb | 0% | 25% | 13% | 0.98 |
| ease-hard | 29% | 0% | 14% | 0.12 |
| ease-worsen | 14% | 0% | 7% | **0.85** |
| elevate-decrease | 0% | 10% | 5% | **0.88** |
| elevate-fall | 18% | 70% | 44% | 0.79 |
| exacerbate-alleviate | 10% | 75% | 43% | **0.87** |
| exacerbate-ease | 0% | 25% | 13% | 0.86 |
| fall-gain | 0% | 8% | 4% | **0.96** |
| fall-increase | 0% | 8% | 4% | 0.56 |
| gain-lose | 33% | 92% | 63% | 0.94 |
| gain-slide | 0% | 8% | 4% | **0.98** |
| increase-decrease | 100% | 100% | 100% | **0.92** |
| increase-drop | 0% | 8% | 4% | 0.58 |
| jump-fall | 29% | 33% | 31% | **0.98** |
| jump-tumble | 0% | 17% | 8% | **0.99** |
| lose-**find** | 17% | 0% | 8% | **0.89** |
| lose-**win** | 17% | 17% | 17% | 0.87 |
| plummet-jump | 50% | 0% | 25% | **0.97** |
| plunge-elevate | 10% | 37% | 23% | 0.77 |
| plunge-**fly** | 20% | 0% | 10% | 0.80 |
| plunge-rise | 10% | 17% | 13% | **0.97** |
| rally-fail | 7% | 0% | 4% | **0.94** |
| rally-retreat | 7% | 42% | 24% | 0.94 |
| rebound-retreat | 0% | 17% | 8% | **0.95** |
| rebound-slip | 20% | 33% | 27% | 0.92 |
| recover-decline | 0% | 29% | 14% | **0.87** |
| recover-lose | 6% | 15% | 11% | 0.84 |
| retreat-advance | 34% | 67% | 50% | **0.92** |
| rise-fall | 73% | 42% | 57% | **0.99** |
| rise-sink | 10% | 25% | 18% | **0.95** |
| sink-elevate | 10% | 0% | 5% | **0.72** |
| sink-**float** | 40% | 0% | 20% | 0.53 |
| slide-climb | 10% | 37% | 23% | **0.99** |
| slide-stable | 0% | 20% | 10% | 0.05 |
| slip-advance | 0% | 17% | 8% | 0.97 |
| slip-slide | 10% | 0% | 5% | **0.99** |
| soar-fall | 43% | 17% | 30% | 0.95 |
| soar-plummet | 0% | 33% | 17% | **0.97** |
| stable-dip | 0% | 8% | 4% | 0.08 |
| stable-**unstable** | 43% | 0% | 21% | **0.85** |
| surge-plunge | 7% | 33% | 20% | **0.98** |
| surge-decrease | 14% | 0% | 7% | 0.76 |
| tumble-climb | 20% | 25% | 23% | **0.98** |
| tumble-rebound | 0% | 33% | 17% | 0.93 |
| volatile-stable | 21% | 69% | 45% | **0.95** |
| volatile-strong | 10% | 0% | 5% | 0.92 |
| weak-strong | 93% | 100% | 96% | **0.96** |
| weak-stable | 0% | 25% | 13% | 0.95 |

## Study 1: People's Antonym Choices

In this study, participants were either given the positive, UP verbs or the negative, DOWN verbs and asked to perform two tasks on the set (a free-generation task, always followed by a match-the-opposite task). The measure was the frequency with which a particular pair was identified in either task.

## Method

**Participants** Twelve students at University College Dublin voluntarily took part in the study; five male and seven female. All were native English speakers. Participants were assigned to one of the two conditions; receiving either all UP verbs or DOWN verbs as prompts in both tasks of the study.

**Materials** The set of UP verbs and DOWN verbs shown in Table 2 were used as the materials.

**Procedure** Participants were given written and verbal instructions indicating that they would be asked to carry out two tasks that involved identifying "the opposites of the presented words". For the *free-generation task* (Task 1) they were read the list of words, one-by-one, and asked to verbally respond to these prompts. Responses were timed and recorded during the study and later transcribed by the experimenter. After Task 1 the experimenter presented the second task. Note there were no constraints on the responses for the first part of the study.

For the *match-the-opposite* task (Task 2), participants were given a sheet of paper with two columns of words. The left column was the list of prompts from the Task 1, and the right column was a list of potential opposites. Their job was to draw lines from the column of prompt-words on the left-hand side to their "best opposite" on the right-hand side. Note, that they were instructed that they could indicate more than one word if they were considered tied for "best opposite". When this task was completed, the sheet was collected and participants were debriefed on the rationale for the study.

**Scoring** Note that whether participants are given the UP or DOWN verb-sets they tend to produce the same pairs; that is, one could be given *rise* and produce *fall*, the *rise-fall* antonym-pair or one could be given *fall* and produce *rise* generating the same *rise-fall* antonym-pair. As there were no clear differences in the pairs identified by participants who were presented either all UP verbs or DOWN verbs, the scoring was performed on the two conditions collapsed together. In scoring the data, we noted the frequency of a particular antonym-pair produced from a particular prompt (e.g., *rise* or *fall*) as a proportion of the total number of presentations of that prompt, in either the first or second task.

## Results & Discussion

*General Characteristics of the Data.* In all, participants identified 114 unique antonym pairs to the 30 presented words (combined UP- and DOWN-verbs). On average, a given prompt-word gave rise to almost five alternative antonym pairs ($M = 4.8$) with a range from 2 (for *weak*, participants produced *weak-strong* and *weak-stable*) or 9 alternative pairs (e.g., *elevate-drop, elevate-fall*). On average, in the *free-generation task* participants suggested one antonym ($M = 1.37$) that was not in the opposing set used in the *match-the-opposite* task (e.g., when presented with *stable* several participants suggested *unstable* as the antonym, but readily chose *volatile* as its antonym in the matching task). Overall, people vary significantly in the antonyms identified for a prompt word. However, for a group of people, there is usually a clear most-frequently-identified antonym. For instance, on average, 96% of participants chose *strong* when prompted with *weak* or *weak* when prompted with *strong*. Table 3 shows the overall percentage for the top two most frequently identified antonym-pairs for each prompt word. Note, that a conservative estimate of chance across both tasks would be close to 5%. This chance-level computation is simply an observation of all available choices in Task 2 along with those free-generation choices in Task 1 that were not available in Task 2. This means that the chance-level estimation of 5% is much more conservative because in Task 1, as the entire English lexicon is available to the participant. Thus, though some percentages are low, they are well above chance.

*The Free-Generation Task.* A notable aspect of the data is how different the percentages are for identified antonyms in the two tasks. The free-generation task allowed participants to name whatever antonym came to mind, some of which were not included in the set for Task 2. However, if one looks at the most-frequently-identified antonyms, there are only five cases (out of 60) where "another" antonym was identified frequently. This means that we can be confident that the match-the-opposite task was not overly constrained in the choices given to participants.

*The Match-the-Opposite Task.* In this task, the choice of antonym was restricted to the 15 contrasting words, with participants being given the option to choose more than one. This is a more constrained task in which to identify antonyms and produced a generally clearer pattern of antonym-pair identification[1]. There are clear winners in terms of favoured antonym pairs; notably, *increase-decrease* (100%), *elevate-fall* (44%) and *alleviate-exacerbate* (43%). Note that some of the low percentages occur because one of the words in the pair is used by another very dominant antonym; so, for example, the listings for *fall-gain* and *fall-increase* are very low (4%; though below chance) because *fall-rise* (implicitly listed in *rise-fall*) has a high percentage (57%).

In itself, this data is interesting but does not answer the posed question of whether these patterns of behaviour are predictable from the argument-distributions of the verbs. In the next study, we turn to this key issue. To reiterate our hypothesis, we expect that an empirical analysis of the distributional similarity between verb-arguments will correlate to the the results of the study presented in this section.

## Study 2: Similarity of Antonym Distributions

Study 1 gives us a set of human data on how people tend to identify antonyms, in this study we compare these identifications to a corpus analysis of the argument distributions of the same words. Our hypothesis was that by taking a distributive approach to knowledge, we might be able to identify antonyms by analysing the arguments they take. Study 1 provides a way of validating our computational analysis of these words' argument distributions.

## Method

**Materials** All the same words used in Study 1 were used in this analysis. We also included the words generated by the participants in Study 1 that were not in our original material list.

**Procedure** Taking the 114 antonym pairs in Study 1, we assembled them into a set of word-vectors by the frequency of their arguments given by Sketch Engine (Kilgarriff, 2004). Each verb had anywhere from 250 to 2,000 arguments in its vector (if a particular word was found in one vector of a pair, but not in the other, it was given a frequency of zero[2]). We examined a number of similarity measures including Euclidean distance, cosine similarity, and Kullback-Leibler divergence. We also compared methods of cutting and smoothing the tails of the distributions to mitigate the effects of low-frequency arguments. Markedly, the most successful measure was cosine similarity, in which the distribution's tail was not cut or smoothed. This measure was applied to the vectors of all words in each of the 114 antonym pairs and similarity scores noted. Correlations were computed between this measure and the proportions for different antonym-pairs in Task 1 and Task 2 separately, as well as the combined totals (see Table 3).

## Results & Discussion

Overall, the argument-distributions of the words provide a moderately effective means for identifying the most-frequently-chosen antonym pairs.

*Correlations to All Antonym-Pairs.* The Pearson correlations between the cosine similarity scores and the proportions in each of the tasks and overall, reveal a moderate correlation ($r = 0.31$) for Task-2 x Cosine-Similarity. The other measures reveal low correlations for Task-1 x Cosine

---

[1] By necessity when a word is generated in Task 1 but not present in Task 2, the percentage has to be 0 in Task 2 (as it was not used as a word prompt).

[2] Note, we also used 1 instead of 0, a technique that is sometimes used to control the effects of the tail of the distribution, but it did not produce notably different results to those reported.

Similarity (*r* = 0.10) and Total-% x Cosine Similarity (*r* = 0.25). It is perhaps not surprising that the best correlation is found in the more constrained task where people's choice of antonym was more restricted. That such correlated regularities could be found for data from a relatively small sample (*n* = 12) is, we believe, very encouraging for the veracity of this technique. However, the correlation only gives us a general sense of the correspondence; the more demanding question is whether the most-frequently-identified antonyms specifically emerge from the computational analysis of argument-distributions.

*Identifying Most-Frequently-Identified Antonyms*. Table 3 shows the top-two most-frequently-identified antonyms for a given prompt word in the UP- and DOWN-verb sets. In the column showing the cosine similar score (*Sim*) for an antonym, when the score is shown in bold it indicates that this was the highest similarity score for all the alternative antonym-pairs in the set. So, in 60% of cases the most-frequently identified antonym-pair was also the one with the highest-similarity score in its set of antonym-pairs. If we widen this assessment to accept the highest and second-highest scored antonym pair, then 87% of the pairs that emerge from the corpus analysis were identified as most-frequent antonyms by participants. This is a very good correspondence between the predictions of the computational measure and the results of the human data.

## General Discussion

Metaphors, and their linguistic instantiations, structure not only the way we converse, but the way we think. In this paper we have shown that a statistical analysis of the argument-distributions can be used to identify antonymic verb-pairs – pairs that refer to opposing metaphors in our knowledge (cf. Lakoff, 1992).

The strongest antonyms identified by participants in Study 1 are shown to be predictable by looking at statistical regularities of word-usage in a corpus. In itself, this is an interesting result, but it also lends support to an emerging body of work on finding meaning behind word-use statistics (see Turney & Pantel, 2010 for a survey). Specifically, vector space models, a form of which we employed in Study 2 of this paper, have been used in computational research on document summarisation, comparison, information extraction, searching, and indexing. These models, have also found cognitive relevance in analogy resolution, semantic priming and comprehension, and word-sense disambiguation. This growing body of work, as well as the current paper, bridges a gap between words and meaning.

In another paper, using the same corpus, we show that metaphoric verbs, exhibit a partially-subsumptive hierarchical structure (Gerow & Keane, 2011b). Both papers show that, in this financial domain, there are clear statistical regularities in word usage that can be used as pointers to the underlying structure and organization of metaphors. We believe that this is an important finding. Indeed, both papers bridge a gap, analogous to the word-meaning gap, between linguistic and conceptual metaphors.

## Acknowledgements
This work was carried out as part of a self-funded MSc in the Cognitive Science programme at the University College Dublin by the first author. Thanks to Trinity College Dublin and Prof. K. Ahmad for support and supervision of the first author's PhD during the preparation of this paper. Thanks to K. Hadfield and four anonymous reviewers for valuable suggestions.

## References
Brown, J. C., Frishkoff, G. A., & Eskenazi, M. (2005). Automatic questions generation for vocabulary assessment. In *Proceedings of the conference on Human Language Technology and Empirical Methods in Natural Language Processing* (pp. 819-826).

Deignan, A. (2005). A corpus-linguistic perspective on the relationship between metonymy and metaphor. *Style, 39 (1)*, 72-91.

Gerow, A. & Keane, M. T. (2011a). Mining the web for the "voice of the herd" to spot stock market bubbles. To appear in *Proceedings of the 22nd International Joint Conference on Artificial Intelligence*.

Gerow, A. & Keane, M. T. (2011b). Identifying metaphoric antonyms in a corpus analysis of finance articles. To appear in *Proceedings of the 33rd Annual Meeting of the Cognitive Science Society*.

Kilgarriff, A., Rychlý, P., Smrž, P., and Tugwell, D. (2004). The sketch engine. In *Proceedings of EU-RALEX* (pp. 105-116).

Lakoff, G. & Johnson, M. (1980). *Metaphors we live by*. Chicago, IL: University of Chicago Press.

Lakoff, G. (1992). The contemporary theory of metaphor. In Andrew, Ortony (ed.) *Metaphor and Thought 2nd Edition*. Cambridge: Cambridge University Press.

Landauer, T. K. & Dumais, S. T. (1991). A solution to Plato's problem: The latent semantic analysis theory of acquisition, induction, and representation of knowledge. *Psychological Review, 104*, 211-240.

Michel, J.-B. et al. (2010). Quantitative analysis of culture using millions of digitized books. ScienceExpress, *10 (1126)*.

Schmid, G. (1994). *TreeTagger ─ a language independent part-of-speech tagger*. (http://www.ims.uni-stuttgart.de/Tools/DecisionTreeTagger.html).

Sekine, S. (1997). *The apple pie parser,* v5.9. (http://nlp.cs.nyu.edu/app/).

Turney, P. D. (2006). Similarity of semantic relations. *Computational Linguistics, 32 (3)*, 379-416.

Turney, P. D. & Pantel, P. (2010). From frequency to meaning: Vector space models of semantics. *Journal of Artificial Intelligence Research,* 37, 141-188

Veale, T. (2004). WordNet sits the SAT: A knowledge-based approach to lexical analogy. In *Proceedings of the 16th European Conference on Artificial Intelligence* (pp. 606-612).